\documentclass{article}

\usepackage[nonatbib, final]{nips_2017}

\usepackage[utf8]{inputenc} 
\usepackage[T1]{fontenc}    
\usepackage{hyperref}       
\usepackage{url}            
\usepackage{booktabs}       
\usepackage{amsfonts}       
\usepackage{nicefrac}       
\usepackage{microtype}      
\usepackage{graphicx}
\usepackage[font=small,labelfont=bf]{caption}
\title{Unsupervised Prostate Cancer Detection on H\&E using Convolutional Adversarial Autoencoders}

\author{
  Wouter~Bulten \\
  Diagnostic Image Analysis Group\\
  Department of Pathology\\
  Radboud University Medical Center\\
  \texttt{wouter.bulten@radboudumc.nl} \\
  \And
  Geert~Litjens \\
  Diagnostic Image Analysis Group\\
  Department of Pathology\\
  Radboud University Medical Center\\
  \texttt{geert.litjens@radboudumc.nl} \\
}

\begin{document}

\maketitle

\begin{abstract}
We propose an unsupervised method using self-clustering convolutional adversarial autoencoders to classify prostate tissue as tumor or non-tumor without any labeled training data. The clustering method is integrated into the training of the autoencoder and requires only little post-processing. Our network trains on hematoxylin and eosin (H\&E) input patches and we tested two different reconstruction targets, H\&E and immunohistochemistry (IHC). We show that antibody-driven feature learning using IHC helps the network to learn relevant features for the clustering task. Our network achieves a F1 score of 0.62 using only a small set of validation labels to assign classes to clusters.

\end{abstract}

\section{Introduction}

The most important prognostic marker for prostate cancer (PCa), the Gleason score \cite{Epstein2010}, is determined by pathologists on H\&E stained specimens and is based on the architectural pattern of epithelial tissue. However, it suffers from high inter- and intra- observer variability. Although supervised deep learning methods are the de facto standard in many medical imaging tasks, unsupervised methods have great potential due to their ability to perform hypothesis-free learning. An unsupervised approach to detecting PCa can potentially find relevant morphological features in the data without relying on a human-engineered grading system such as the Gleason score. 

\begin{figure}[h]
\vspace{-0.3cm}
\centering
\includegraphics[width=\textwidth]{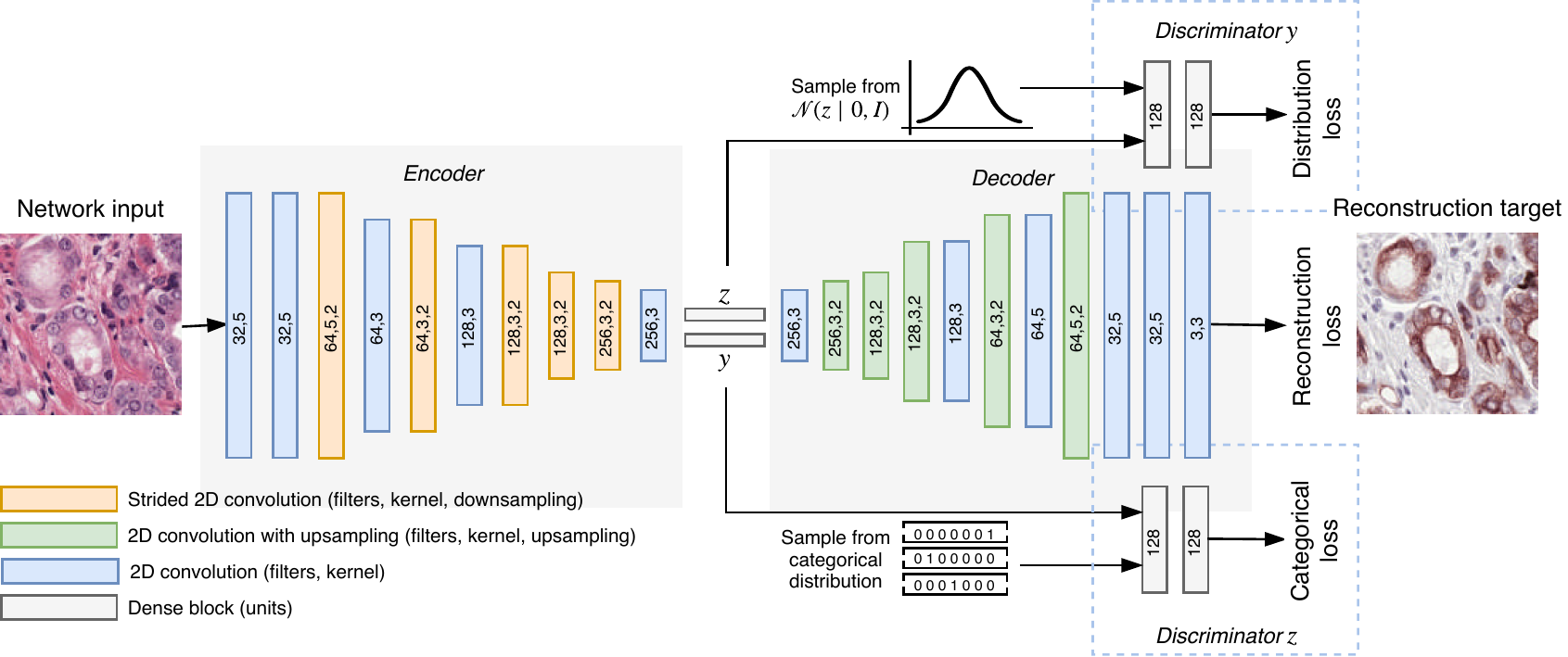}
\caption{Overview of the clustering adversarial autoencoder (CAAE). By reconstructing an IHC patch from a H\&E patch through a style vector, $z$, and a cluster vector, $y$, the network learns to cluster prostate tissue. This antibody-driven feature learning forces the network to learn more relevant encodings.}
\label{fig:network}
\vspace{-0.25cm}
\end{figure}

We propose a new unsupervised method based on adversarial autoencoders \cite{Makhzani2016} to cluster tissue in tumor and non-tumor that does not rely on labels during training: a clustering adversarial autoencoder (CAAE). As opposed to normal autoencoders, our CAAE clusters tissue as part of the training process and does not require post-processing in terms of kMeans, t-SNE or other clustering methods.

\section{Methods}

\textbf{Data description.} We trained our CAAE on patches extracted from 94 registered whole slide image (WSI) pairs, divided into training (54) and test (40), from patients that underwent a radical prostatectomy at the  Radboud University Medical Center. Each pair consists of a H\&E slide, and a slide that was processed using IHC with an epithelial (CK8/18) and basal cell (P63) marker. In the IHC, prostate cancer can be identified as areas where the epithelial marker is present, but the basal cell marker is absent. We randomly sampled 108,000 patches with a size of $(128 \times 128)$ pixels at 5x magnification, sampling more from regions with epithelial tissue (using \cite{Bulten2018}). We did not explicitly sample from tumor regions.

\begin{figure} [h!]
\centering
\footnotesize
\begin{minipage}[t]{.35\textwidth}
\centering
  \includegraphics[height=5.1cm]{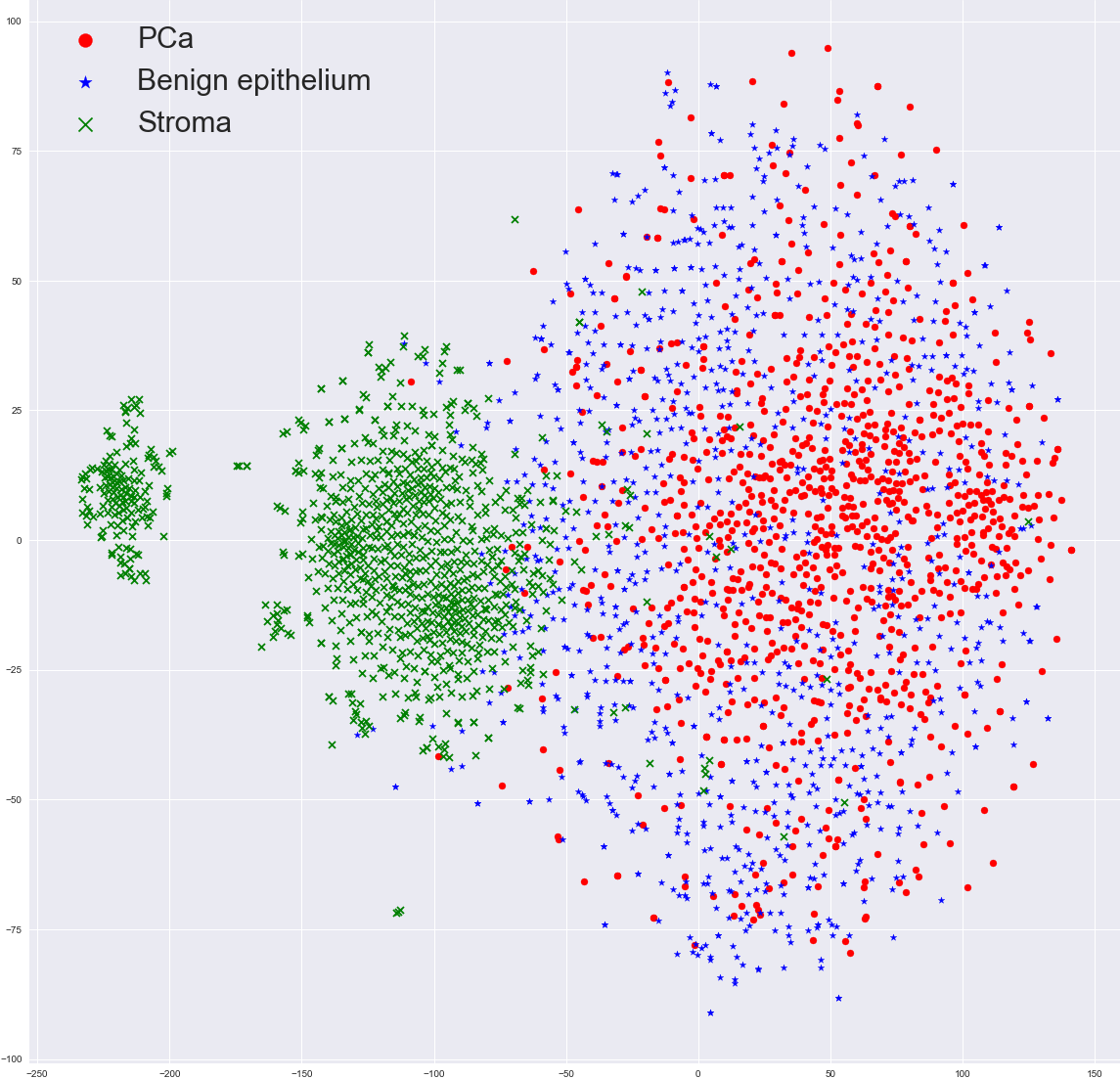}
  \caption{  \label{fig:tsne} t-SNE of normal adversarial autoencoder (latent space = 100) trained and applied to the test set showing that the data is separable.}
  \label{fig:test1}
\end{minipage}\hspace{0.09\textwidth}
\begin{minipage}[t]{.55\textwidth}
  \centering
   \includegraphics[height=5.1cm]{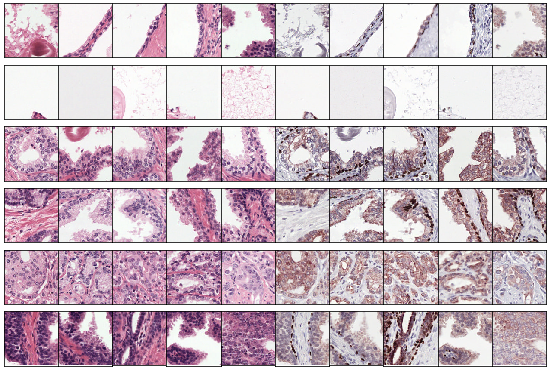}
  \caption{\label{fig:test2} Patches that maximize certain clusters (rows). Some clusters capture a class perfectly, e.g. stroma in row 1 and 2 and tumor in row 5. Some clusters look similar but contain both benign epithelium and tumor (row 6).}
  
\end{minipage}
\vspace{-0.5cm}
\end{figure}

\textbf{Antibody-driven feature learning.} The H\&E patches were used as training input. We tested two reconstruction targets: using the input patch as the target (H\&E to H\&E) and using the IHC version of the same patch (H\&E to IHC). By using a IHC patch as reconstruction target we force the network to learn which features in H\&E correspond to features in the IHC; we hypothesized that this antibody-driven feature learning results in more relevant encodings as better information for tumor/non-tumor separation is present in the IHC. As an initial experiment, we trained a regular adversarial autoencoder on the H\&E to IHC mapping and performed \textit{t-SNE} to see whether the data is separable (Figure \ref{fig:tsne}).

\begin{figure}[h]
\centering
\includegraphics[width=\textwidth]{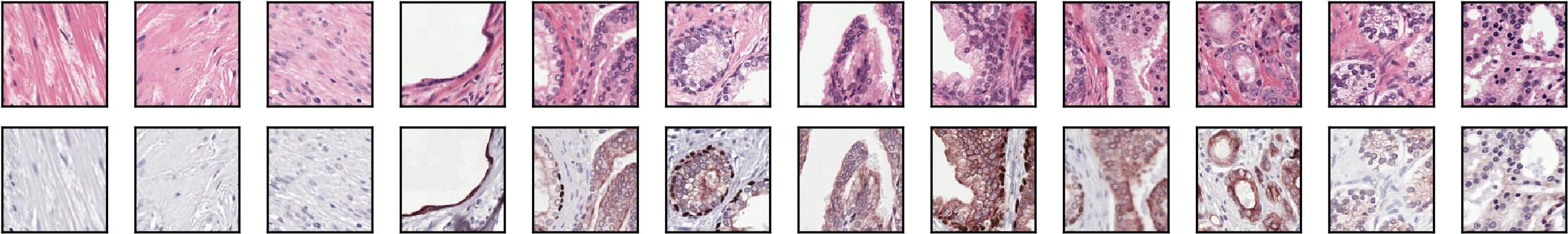}
\caption{Example patches (first row H\&E, second IHC). Column 1-4 show stroma, 5-8 benign epithelium, 9-12 various stages of PCa. Differences between benign tissue and PCa are subtle, especially in H\&E.}
\label{fig_sim}
\end{figure}
\textbf{Network architecture.} Our CAAE consists of four subnetworks (Figure \ref{fig:network}) and two latent vectors as the embedding. The first latent vector, $y$ (size 50), represents the cluster vector and is regularized by a discriminator to follow a one-hot encoding. The second latent vector, $z$ (size 20), represents the style of the input patch following a Gaussian distribution. Training the CAAE forces the network to describe high level information in the cluster vector using one of the 50 classes, and low-level reconstruction information in the style vector. The ratio between the length of the two vectors is critical, a too large $z$ and the network will encode all information using the more easy to encode style vector and disregard the cluster vector.

The CAAE is trained in three stages on each minibatch. First, the autoencoder itself is updated to minimize the reconstruction loss. Second, both discriminators are updated to regularize $y$ and $z$ using data from the encoder and the target distribution. Last, the adversarial part is trained by connecting the encoder to the two discriminators separately and maximizing the individual discriminator loss, updating only the encoder's weights. This forces the latent spaces to follow the target distribution.

\textbf{Validation.} We sampled patches from the test slides from three areas: 1) 1000 stroma (connective tissue) or patches without epithelium; 2) 1000 benign epithelium (healthy tissue); and 3) 2000 PCa (tumor tissue) patches. The tumor patches are not perfect representatives of their class; e.g. a tumor patch can also contain benign epithelium due to the coarseness of the reference annotations. All patches are passed through the encoder to retrieve their embeddings. We took 200 patches of each class to map clusters to class labels, all other patches in the test set are assigned using this mapping. We also computed the scores using all patches to optimize the mapping.

\begin{figure}[h!]
\centering
\includegraphics[width=\textwidth]{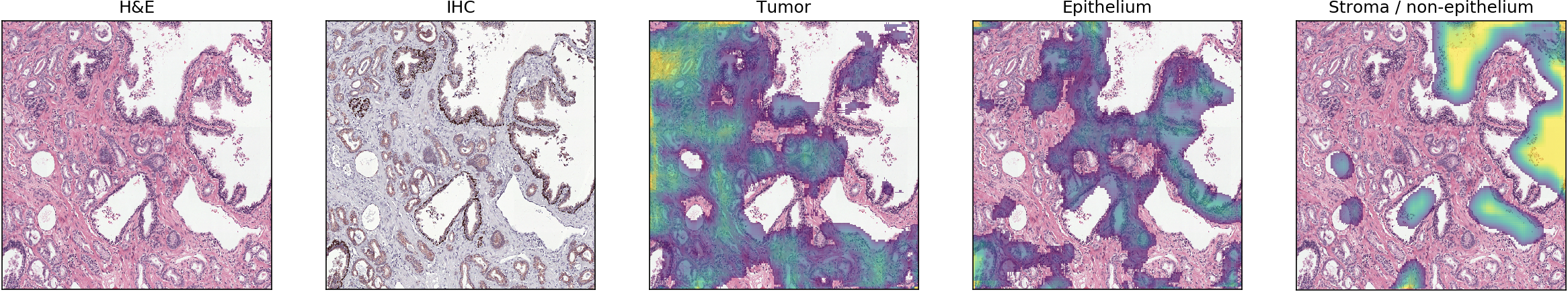}
\caption{Encoder applied as a sliding window to a larger area. Overlay shows the relative frequency that each pixel was assigned a certain class.}
\label{fig:overlay}
\end{figure}

\begin{table}[h!]
\centering
\caption{Classification performance of the networks for tumor vs non-tumor and individual classes.}
\label{tab:results}
\begin{tabular}{@{}lllll|l@{}}
\toprule
                                & \multicolumn{3}{l}{\textbf{Tumor, non-tumor}}            &             & \textbf{Stroma, benign, tumor} \\
                                & \textbf{Accuracy} & \textbf{Precision} & \textbf{Recall} & \textbf{F1} & \textbf{Accuracy}             \\ \midrule
H\&E to H\&E                    & 59\%              & 0.622              & 0.448           & 0.520       & 52\%                           \\
H\&E to H\&E (using all labels) & 63\%              & 0.588              & 0.873           & 0.703       & 60\%                           \\
H\&E to IHC                     & 68\%              & 0.739              & 0.544           & 0.621       & 63\%                           \\
H\&E to IHC (using all labels)  & 75\%              & 0.715              & 0.833           & 0.769       & 72\%                           \\ \bottomrule
\end{tabular}
\end{table}

\section{Results \& Discussion}

Our CAAE, trained to reconstruct IHC from H\&E, achieves a F1 score of 0.62 in discriminating tumor versus non-tumor (Table \ref{tab:results}). We also observe that the clusters represent distinct features of the input data (Figure \ref{fig:overlay}). In comparison, the H\&E to H\&E network performs far worse with a F1 in tumor versus non-tumor of 0.52, showing that, although the reconstruction task is more difficult, there is added benefit in learning the cross-domain mapping. To show the maximum performance of our network we computed the F1 score using all available labels, resulting in a score of 0.77.

While these results leave enough room for improvement, our network achieves these scores without using any labeled training data on a very heterogeneous and noisy dataset. Pathologists grade PCa on multiple levels and use larger field of views than the patches in our dataset. A logical next step would be to increase the field of view of the autoencoder.

\footnotesize
\bibliographystyle{ieeetr}
\bibliography{./references}

\end{document}